\documentclass[conference]{IEEEtran}
\IEEEoverridecommandlockouts

\usepackage{cite}
\usepackage{amsmath,amssymb,amsfonts}
\usepackage{algorithmic}
\usepackage{graphicx}
\usepackage{textcomp}
\usepackage{xcolor}
\usepackage[numbers]{natbib}
\usepackage{etoolbox}
\def\BibTeX{{\rm B\kern-.05em{\sc i\kern-.025em b}\kern-.08em
    T\kern-.1667em\lower.7ex\hbox{E}\kern-.125emX}}

\makeatletter

\def\ps@IEEEtitlepagestyle{
  \def\@oddfoot{\mycopyrightnotice}
  \def\@evenfoot{}
}
\def\mycopyrightnotice{
  {\footnotesize 978-0-7381-2333-2/20/\$31.00~\copyright~2020 IEEE\hfill} 
  \gdef\mycopyrightnotice{}
}

\@ifundefined{showcaptionsetup}{}{
 \PassOptionsToPackage{caption=false}{subfig}}
\usepackage{subfig}
\makeatother

\usepackage{eso-pic}
\newcommand\AtPageUpperMyright[1]{\AtPageUpperLeft{
 \put(\LenToUnit{0.5\paperwidth},\LenToUnit{-1cm}){
     \parbox{0.5\textwidth}{\raggedleft\fontsize{9}{11}\selectfont #1}}
 }}
\newcommand{\conf}[1]{
\AddToShipoutPictureBG*{
\AtPageUpperMyright{#1}
}
}
\begin{document}

\title{Sentiment analysis in Bengali via transfer learning using multi-lingual BERT\\}
\conf{2020 23\textsuperscript{rd} International Conference on Computer and Information Technology (ICCIT), 19-21 December, 2020.}

\author{\IEEEauthorblockN{Khondoker Ittehadul Islam}
\IEEEauthorblockA{\textit{Computer
Science and Engineering
} \\
Shahjalal University of
Science 
\\and Technology 
\\Sylhet, Bangladesh \\
shanislam6@gmail.com}
\and
\IEEEauthorblockN{ Md Saiful Islam}
\IEEEauthorblockA{\textit{Computer
Science and Engineering
} \\
Shahjalal University of
Science
\\and Technology
\\Sylhet, Bangladesh \\
saiful-cse@sust.edu}
\and
\IEEEauthorblockN{Md Ruhul Amin}
\IEEEauthorblockA{\textit{Computer and Information Science} \\
Fordham University\\
New York, USA \\
mamin17@fordham.edu}
}

\maketitle

\begin{abstract}
Sentiment analysis (SA) in Bengali is challenging due to this Indo-Aryan language's highly inflected properties with more than 160 different inflected forms for verbs and 36 different forms for noun and 24 different forms for pronouns. The lack of standard labeled datasets in the Bengali domain makes the task of SA even harder. In this paper, we present manually tagged 2-class and 3-class SA datasets in Bengali. We also demonstrate that the multi-lingual BERT model with relevant extensions can be trained via the approach of transfer learning over those novel datasets to improve the state-of-the-art performance in sentiment classification tasks. This deep learning model achieves an accuracy of 71\% for 2-class sentiment classification compared to the current state-of-the-art accuracy of 68\%. We also present the very first Bengali SA classifier for the 3-class manually tagged dataset, and our proposed model achieves an accuracy of 60\%. We further use this model to analyze the sentiment of public comments in the online daily newspaper. Our analysis shows that people post negative comments for political or sports news more often, while the religious article comments represent positive sentiment. The dataset and code is publicly available \footnote{ https://github.com/KhondokerIslam/Bengali\_Sentiment}.
\end{abstract}


\section{Introduction}
Sentiment classification is the task of analyzing a piece of text to predict the orientation of the attitude towards an event or opinion. The sentiment of a text can be either positive or negative. Sometimes, a neutral perspective is also considered for classification. SA has many different applications, such as reducing the early age suicide rate by identifying cyberbullying \cite{nahar2012sentiment}, discouraging unwarranted activities towards a particular community through hate-speech detection \cite{badjatiya2017deep}, and monitoring public response towards a proposed government bill \cite{hurlimann2016twitter} among many others.

The task of SA has achieved superior improvement in other languages, i.e. English - about 97.1\% accuracy for 2-class \cite{raffel2019exploring} and 91.4\% accuracy for 3-class SA \cite{hong2019multi}. But only a few research works have been published for the SA in Bengali. This is because we lack quality datasets in Bengali for training a computation model for the sentiment classification. However, in the last few years, we have seen the rise of Internet users in the Bengali domain mostly due to the development of wireless network infrastructure throughout South East Asia. This resulted in a massive increase in the total number of online social network users as well as newspaper readers. So it became comparatively easier to collect the public comments posted online on the Bengali news websites. 

\begin{table}
\caption{SA of public comment published in the online newspaper. We collected 334 comments for each of the politics, sports, and religion categories. We only collected one comment from a randomly selected news article. In the table, we present the percentage of the total comments classified into three different sentiment classes.}
\label{tab:Cate_senti}
\centering
 \begin{tabular}{|l | c | c | c | } 
 \hline
  & Negative & Neutral & Positive  \\ [0.5ex] 
 \hline
Politics &
66\% &
24\% &
10\% \\
 \hline
 
Sports &
52\% &
38\% &
10\% \\
 \hline

Religion &
42\% &
8\% &
50\% \\
 \hline
\end{tabular}
\end{table}

Thus we created two SA datasets for 2-class and 3-class SA in Bengali and trained a multi-lingual BERT model via transfer learning approach for sentiment classification in Bengali, referred as $BERT_{BSA}$ in this paper. $BERT_{BSA}$ achieves an accuracy of 71\% for the 2-class and 60\% for the 3-class manually tagged dataset. We further use this model to analyze the sentiment of 1,002 public comments collected from the online daily newspaper. Table \ref{tab:Cate_senti} shows that in general, sentiment in public comments is positive for religious news articles, while that is negative for political or sports news articles. In this paper, we present the following contributions:
\begin{itemize}
    \item We created two datasets for SA in Bengali and made it public for further research work. We discuss the methodology we used to create the datasets in the Section \ref{dataset}. 
    \item We introduce a deep learning model for SA in Bengali, $BERT_{BSA}$, that performs better compared to other existing models that are trained with word2vec or fastText embedding. We discuss the model and in Section \ref{method}.
    \item We evaluate $BERT_{BSA}$ and compare it to other models trained with Word2Vec and fastText embeddigns using the 2-class and 3-class Bengali SA datasets. We discuss the results in the Section \ref{result}.
    \item We conduct experiments to investigate application level use of Bengali SA on newspaper comments in three aspects, such as politics, sports and religion, and show that public sentiment is biased towards positive polarity for the news articles related to religion.
\end{itemize} 


\section{Related Work} \label{related}

Bidirectional Encoder Representations from Transformers, or BERT \cite{devlin2018bert}, is an unsupervised language representation model that had been pre-trained using large plain text corpus. BERT makes use of transformer, an attention mechanism to learn the contextual relations between words. BERT is fundamentally different from the context-free models such as Word2Vec or GloVe that generate a single word embedding representation for each word in the vocabulary \cite{mikolov2013efficient}. Instead, BERT takes into account the context for each occurrence of a given word in a sentence. For instance, the vector for ``running" will have the same Word2Vec or GloVe vector representation for both of its occurrences in the sentences ``He is running a company" and ``He is running a marathon." But BERT will provide two contextualized embedding vectors based on the appearance of "running" in two different sentences.

BERT is very popular for aspect-based sentiment analysis by either fine-tuning BERT's pre-trained model \cite{sun2019utilizing, karimi2020adversarial, rietzler2019adapt, xu2020target} or using the benchmark dataset for question-answering \cite{xu2019bert}. However, in the pre-BERT era, research works used other end-to-end deep network layers like LSTM, BiLSTM, CNN, etc. Lei et al. \cite{lei2018multi} integrated with three kinds of sentiment linguistic knowledge (e.g., sentiment lexicon, negation words, intensity words) into the deep neural network via attention mechanisms. In another research work, Baziotis et al. \cite{baziotis2017datastories} used LSTM networks augmented with two kinds of attention mechanisms, on top of pre-trained word embedding for sentiment classification and achieved the rank $1^{st}$ (tie) at the  SemEval-2017 Task 4 Subtask A \cite{nakov2019semeval}.

In spite of such advances in English SA, only a few notable works were done on Bengali SA. Sharfuddin et al. \cite{sharfuddin2018deep} use term frequency–inverse document frequency (tf-idf) and BiLSTM to predict the sentiment of unseen sentences accurately and holds the current state-of-the-art performance on 2-class Bengali sentiment classification in a small balanced dataset. On the other hand, Karim et al. \cite{karim2020classification} focused primarily on building a Bengali word embedding which was incorporated into a Multichannel Convolutional LSTM (MConv-LSTM) network for predicting different types of tasks including sentiment analysis. 

The lack of quality datasets and complex linguistics feature of Bengali language make the task of SA very challenging. In this research work, we contribute two manually tagged datasets for 2-class and 3-class sentiment classification in Bengali. We trained our proposed model $BERT_{BSA}$ as well as the model proposed by Sharfuddin et al. \cite{sharfuddin2018deep} on those datasets and compare the performance of both the models in the section 6.   


\section{Dataset for Bengali SA} \label{dataset}


We choose Prothom Alo\footnote{https://www.prothomalo.com/}, an online news portal, for collecting user's comment. We selected a total of 10 popular newspaper topics (Table \ref{tab:Topic_Dis}) and scrapped a total of 40,354 comments. Upon filtering out noisy comments, we tagged each opinion in to one of three sentiments: \textit{Negative}, \textit{Neutral}, or \textit{Positive} by three independent individuals. Our final dataset contains 17,852 entries (Table \ref{tab:Topic_Dis}). Each of those entries and corresponding tags were validated by an expert Bengali linguistics. In order to analyze our findings and compare with the current state-of-the-art performing model, we made the dataset suitable for 2-class classification tasks by removing the neutral class resulting in a total of 13,120 entries. We present the distribution of the 3-class dataset across training, validation test sets in the Table \ref{tab:DatasetTable}. Moreover, detailed statistics of our final corpus is presented in the Table \ref{tab:StatTable}

\begin{table}
\caption{Distribution of total annotated sample across 10 topics.}
\label{tab:Topic_Dis}
\centering

 \begin{tabular}{|l | c | } 
 \hline
 Topics & No. of data   \\ [0.5ex] 
 \hline
 Sports & 2,332\\
 \hline
 Economy & 1,759 \\
 \hline
 Entertainment & 2,697 \\
 \hline
 
 International & 1,985 \\
 \hline
 
 Education & 1,956\\
 \hline
 
 Technology & 1,282\\
 \hline

 Lifestyle & 1,803\\
 \hline

 Fashion & 1,108\\
 \hline

 Food & 1,343\\
 \hline

 Travel & 1,587\\ [1ex] 
 \hline

 \hline\hline
 Total & 17,852 \\
 \hline
 
\end{tabular}

\end{table}

\begin{table}
\caption{ Distribution of sample across training, validation and test sets.}
\label{tab:DatasetTable}
\centering

 \begin{tabular}{|l | c | c | c | } 
 \hline
  & Train & Valid & Test  \\ [0.5ex] 
 \hline
 Negative & 6011 & 1060 & 1280 \\
 \hline
 Neutral & 3277 & 578 & 877 \\
 \hline
 Positive & 3338 & 588 & 843 \\ [1ex] 
 \hline\hline
 Total & 12626 & 2226 & 3000 \\
 \hline
 
\end{tabular}

\end{table}

\section{Methodology} \label{method}

\begin{figure}
    \centering
    \includegraphics[width=8cm]{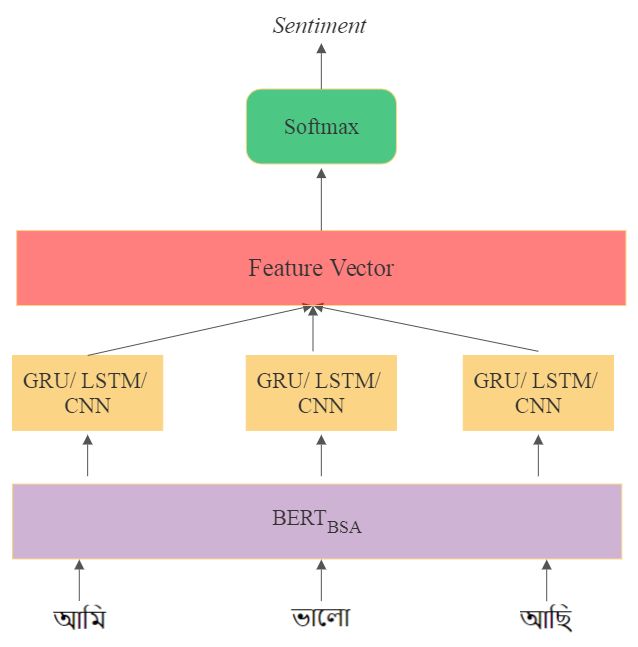}
    \caption{Sentiment analysis in Bengali via transfer learning using multi-lingual BERT. In this figure we present the sentiment classification of a Bengali text which can be translated to "I am doing fine" in English. }
    \label{fig:model_archi}
\end{figure}

In this section, we present the implementation details of our experimental setup. We used the multilingual BERT, \textit{bert-base-multilingual-cased}\footnote{https://github.com/google-research/bert/blob/master/multilingual.md}, as it is the only model that was trained with Bengali corpus up until now. We extended the model with three different end-to-end deep network layers: Gated Recurrent Unit (GRU) \cite{cho2014learning}, Long Short Term Memory (LSTM) \cite{hochreiter1997long}, and Convolutional Neural Network (CNN) \cite{fukushima1988neocognitron}. We performed three experiments with the three different BERT extensions. The architecture of our proposed model $BERT_{BSA}$ is depicted in the Figure \ref{fig:model_archi}. BERT produces contextualized embedding vector for each word which are passed through one of the three deep network layers: GRU, LSTM, or CNN, in the different experiment. The output neurons for each word from the intermediate layer is concatenated to form the feature vector. Finally, this vector is passed through fully connected neural dense layer for dimension reduction. The final reduced vector is passed through softmax for sentiment classification. The results are presented in Table \ref{tab:2-class_result} for 2-class and 3-class classification.

We also implemented two other deep learning architectures with pre-trained word embeddings, Word2Vec and fastText which were extended with GRU, LSTM, and CNN deep network layers. Word2Vec  \cite{mikolov2013efficient} has been quite successful for SA across several languages, including Bengali \cite{al2017sentiment}. Also, fastText \cite{bojanowski2017enriching} gained huge popularity in Bengali text analysis mainly due its operation on character level n-grams \cite{ritu2018performance,khatun2019authorship,roy2019unsupervised}. So we compare the performance of SA for the 2-class and 3-class classification of these models with that of $BERT_{BSA}$, and present the outcomes in Table \ref{tab:2-class_result}.

\begin{table}
\caption{Data Statistics.}
\label{tab:StatTable}
\centering

 \begin{tabular}{|l | c | } 
 \hline
  & Words   \\ [0.5ex] 
 \hline
 Longest Sentence & 128  \\
 \hline
 Average Sentence Length & 45  \\
 \hline
 Total Words & 312569 \\
 \hline
 Non-Bengali Words & 0 \\
 \hline
 
\end{tabular}

\end{table}

For three different deep learning architectures with $BERT_{BSA}$, Word2Vec, and fastText, we used the following model parameters: 
\begin{itemize}
    \item \textbf{GRU}: \indent For 2-class classification tasks, we used single bidirectional GRU layer where the final layer outputs 300 neurons per word with a dropout of 0.5. For 3 class, we had two bidirectional GRU layers where the final layer outputs 350 neurons per word with a dropout of 0.5.
    \item \textbf{LSTM} \indent For 2-class classification tasks, we used 3 bidirectional LSTM layers with an output of 100 neurons per cell and dropout of 0.5. For 3-class classification task, we used and output of 512 neurons per word using only a single bidirectional LSTM layer. 
    \item \textbf{CNN} \indent For 2-class classification tasks, we used a CNN layer with kernel size of [3, 3] and filter size of [64, 100]. For 3-class classification task, we used a single CNN layer with kernel size of [1, 2, 3, 4] and filter size of [200].
\end{itemize}
For all the models, Adam was used as an optimizer and \textit{L2} was used for regularization.





\section{Result Discussion} \label{result}

The outcome of all the different experiments that we performed are presented in the Table \ref{tab:2-class_result}. As the dataset is based on public opinion, most of the words are informal.  Therefore,  BERT's ability to manage out-of-vocabulary words effectively helped RNN architecture to carry meaningful context over a long period of time. These resulted RNN architecture performing better with BERT. However, only for 2-class classification task, CNN performed better with fastText over BERT and Word2Vec. Moreover, amongst the RNN architectures GRU performed better than LSTM due to the small-scale dataset \cite{chung2014empirical}.

\begin{table}
\caption{ Sentiment classification accuracy of Word2Vec, fastText and $BERT_{BSA}$ for three different extensions. $BERT_{BSA}$ produced the best accuracy for both the 2-class and 3-class SA tasks.}
\label{tab:2-class_result}

\centering
 \begin{tabular}{|l | c | c | c | c | } 
 \hline
  &  & GRU & LSTM & CNN  \\ [0.5ex] 
 \hline


& Word2Vec &
0.67 &
\textbf{0.68} &
0.66 \\
2-class & fastText & 
0.68 &
0.68 &
\textbf{0.69}  \\
& BERT\textsubscript{BSA} & 
\textbf{0.71} &
0.70 & 
0.67 \\

\hline

& Word2Vec &
\textbf{0.57} &
0.54 &
0.55 \\
3-class & fastText & 
\textbf{0.58} &
\textbf{0.58} &
0.56  \\
& BERT\textsubscript{BSA} & 
\textbf{0.60} &
0.59 & 
0.58 \\

\hline

\end{tabular}

\end{table}

Furthermore, we ran state-of-the-art model proposed by Sharfuddin et al. \cite{sharfuddin2018deep} with our 2-class dataset in our environment and got an accuracy of 68\% whereas our model resulted in 71\% accuracy. $BERT_{BSA}$ is the only model in Bengali that performed a 3-class classification and resulted in 60\% accuracy. This verifies that $BERT_{BSA}$ model with GRU substantially beats the state-of-the-art model that uses tf-idf vectorization with a BiLSTM architecture.










 


As our ultimate goal is to use this model in real-life applications, we scrapped user's comment from popular Bengali newspaper sites from three different topics: sports, religion and politics. A total of 1,002 comment had been scraped with 334 comments for each topic from contents ranging from January 2020 to April 2020. Table \ref{tab:Cate_senti} shows some interesting findings of native Bengali speaker. Towards politics, Bengali people seems to be much critical. With around 65\% of comments seems negative, it clearly states that people are not happy on the ongoing politics taking place in this region. Moreover, people speaking this language tends to criticize their own sport half of the time with percentage of applaudable comment is equal to that politics. Moreover, with neutral being close to negativity, it can be claimed that Bengali people wants better result in sports. However, having the most sentiment as positive in religious topics defines Bengali people are respectful towards diversity and also being faithful believer.

\section{Conclusion}

In this paper, we presented two manually tagged novel datasets for SA in Bengali. We also introduced \textit{BERT\textsubscript{BSA}}, a deep learning model for SA in Bengali, which outperforms all other models. We achieved state-of-the-art performance for both the 2-class and 3-class SA tasks in Bengali. Moreover, we took a step closer to apply SA model to a real world application by analyzing public sentiment on newspaper topics. The result shows that for religious news comments people tend to possess a positive sentiment whereas for political and sports news comments, people possess negative sentiment. However, this research is a work in progress and will be regularly updated with new insights. We are continuing to increase the size of SA datasets in Bengali and we will explore the application of other deep learning models for better results. We hope that the improved performance of SA in multi-class classification tasks presented in this paper will help many ground-breaking applications like cyberbullying identification as well as hate-speech detection in Bengali.

\section{Acknowledgement}

We would like to thank Shahjalal University of Science and Technology (SUST) and SUST NLP research group for their support.






\renewcommand\bibname{References}
\bibliographystyle{IEEEtran}
\bibliography{Bibliography.bib}

\end{document}